\documentclass[letterpaper]{article} % DO NOT CHANGE THIS
\usepackage{aaai24}  % DO NOT CHANGE THIS
\usepackage{times}  % DO NOT CHANGE THIS
\usepackage{helvet}  % DO NOT CHANGE THIS
\usepackage{courier}  % DO NOT CHANGE THIS
\usepackage[hyphens]{url}  % DO NOT CHANGE THIS
\usepackage{graphicx} % DO NOT CHANGE THIS
\usepackage{natbib}  % DO NOT CHANGE THIS AND DO NOT ADD ANY OPTIONS TO IT
\usepackage{caption} % DO NOT CHANGE THIS AND DO NOT ADD ANY OPTIONS TO IT
\usepackage{algorithm}
\usepackage{algorithmic}
\usepackage{subcaption}
\usepackage{booktabs}
\usepackage{array,multirow,graphicx}
\usepackage{amssymb}% http://ctan.org/pkg/amssymb
\usepackage{pifont}% http://ctan.org/pkg/pifont\
\newcommand{\xmark}{{\ding{55}}}%

\usepackage{colortbl}
\usepackage[symbol]{footmisc}

\usepackage{todonotes}

\usepackage{newfloat}
\usepackage{listings}
\usepackage{amsmath}
\DeclareCaptionStyle{ruled}{labelfont=normalfont,labelsep=colon,strut=off} % DO NOT CHANGE THIS
\floatstyle{ruled}
\newfloat{listing}{tb}{lst}{}
\floatname{listing}{Listing}
\title{MIND: Multi-Task Incremental Network Distillation}
\author{
    Jacopo Bonato\equalcontrib \textsuperscript{1}, Francesco Pelosin\equalcontrib \textsuperscript{1} \textsuperscript{2} \thanks{Work done while at Leonardo Labs}, Luigi Sabetta\equalcontrib \textsuperscript{1}, Alessandro Nicolosi \textsuperscript{1}\\% \thanks{wsss},\\
}
\affiliations{
    \textsuperscript{1}Leonardo Labs, Rome, Italy\\
    \textsuperscript{2}Covision Lab, Brixen South-Tyrol, Italy\\
    \tt\small{\{jacopo.bonato, luigi.sabetta\}.ext@leonardo.com}\\
    \tt\small{francesco.pelosin@covisionlab.com}\\
\tt\small{alessandro.nicolosi@leonardo.com}
}

\begin{document}

\maketitle

\begin{abstract}

%%%%%%%%%%%%%%
The recent surge of pervasive devices that generate dynamic data streams has underscored the necessity for learning systems to adapt continually to data distributional shifts. To tackle this challenge, the research community has put forth a spectrum of methodologies, including the demanding pursuit of class-incremental learning without replay data. In this study, we present MIND, a parameter isolation method that aims to significantly enhance the performance of replay-free solutions and achieve state-of-the-art results on several widely studied datasets. Our approach introduces two main contributions: two alternative distillation procedures that significantly improve the efficiency of MIND increasing the accumulated knowledge of each sub-network, and the optimization of the BachNorm layers across tasks inside the sub-networks. Overall, MIND outperforms all the state-of-the-art methods for rehearsal-free Class-Incremental learning (with an increment in classification accuracy of approx. $+6\%$ on CIFAR-100/10 and $+10\%$ on TinyImageNet/10) reaching up to approx. $+40\%$ accuracy in Domain-Incremental scenarios. Moreover, we ablated each contribution to demonstrate its impact on performance improvement. Our results showcase the superior performance of MIND indicating its potential for addressing the challenges posed by Class-incremental and Domain-Incremental learning in resource-constrained environments.

% The code and experiments are publicly available at the following link [insert link].
\end{abstract}

\section{Introduction}

Despite the remarkable achievements witnessed in deep learning in recent years, a fundamental challenge that remains unresolved pertains to enabling lifelong learning in deep neural networks. Learning continually would unlock the ability of artificial networks to learn new tasks sequentially by adapting to distributional shifts and overcoming the so-called catastrophic forgetting. This issue arises because the network's parameters adapted to the incoming task become ill-suited for the old data, resulting in performance degradation over time. Mitigating catastrophic forgetting typically involves retraining the system from scratch using both old and new data \cite{survey2, survey_masana}, which is not only expensive but also fails to adapt to future scenarios automatically. In response, numerous approaches have been proposed. Some methods employ replay buffers to approximate the old data while training on new data \cite{gdumb, er, ermir}. In contrast, others instantiate new parameters as new tasks emerge \cite{pnn, dytox}. Additional approaches leverage regularization techniques in the parameter space to address this issue \cite{ewc, si}.

Recently, the deep learning research community has emphasized the importance of exploring compositional approaches in learning systems. The compositional nature of networks has been identified as a crucial aspect of intelligent systems, wherein each sub-module can be selectively accessed to solve specific tasks. For example, the research on LLMs demonstrates through multimodal pipelines how such approaches are effective and efficient: sub-modules are tailored for a particular task characterized by different modalities (i.e. LLAVA \cite{llava} LENS \cite{LENS} and Flamingo \cite{flamingo}). For this reason, systems that exhibit compositionality offer the advantage of being compact, enabling them to address multiple tasks within a single architecture while minimizing memory requirements. In particular, this modular structure of intelligent systems is aligned with some neuroscientific theories like the complementary learning system (CLS) theory \cite{cls} which describes how the brain employs two distinct and specialized systems for learning and memory. Among this research stream, the continual learning community proposed several architectures characterized by a slow- and a fast-learner coupled to tackle incremental learning tasks\cite{fast_slow}.

Along this research line, we propose a new method called MIND that belongs to the category of parameter isolation approaches \cite{survey_masana}, where sub-regions of the network, called sub-networks, are allocated for tackling individual tasks. However, these sub-regions are not completely disjoint between each other and share a fraction of parameters facilitating the transfer of previously acquired knowledge for future task-solving. In this context, MIND exploits a distillation procedure~\cite{distillation} to encapsulate and compress the knowledge from a new model trained for each new task into a sub-network fragment. Furthermore, we propose a variation of the optimization procedure of MIND that works under memory limitation, involving a self-distillation procedure where the new model is substituted by MIND itself. Altogether MIND significantly enhances the performance of standard parameter isolation approaches like PackNet~\cite{packnet} and demonstrates superior effectiveness in learning new data while retaining past information compared to the current state-of-the-art methods. We make code and experiments available at \textcolor{cyan}{https://github.com/Lsabetta/MIND}.

In summary, the contributions of our work can be outlined as follows:

\begin{itemize}

    \item We develop a novel parameter isolation approach equipped with a distillation mechanism. This optimization procedure makes use of the knowledge acquired by a new model trained for each new task and transfers it into a sub-network fragment of MIND by matching output probability distributions of the new model and a sub-network of MIND. Importantly, starting from this procedure we propose a different distillation mechanism where MIND self-distills its knowledge about a task inside a single sub-network. Hence, this approach allows MIND with self-distillation to work under memory limitation.

    \item We propose different policies to select sub-networks for each task. In particular, when using a new model we select randomly the sub-network weights of MIND whereas when performing self-distillation we can select the weights with the highest absolute value.

    \item As an integrated part of our method, we introduced a gating mechanism to be applied in our backbone. The gating mechanism guides the gradient flow during backpropagation and approximates more correctly its computation. 
    %\item  During inference we propose a simple but effective post-hoc selection mechanism to select the correct sub-network and the correct output class.
    
    \item We provide a broader and solid experimental framework by testing MIND in 4 different datasets in Class Incremental (CI) scenario (i.e. new classes are presented for each new task).  MIND shows optimal performance and outperforms the state-of-the-art methods. Moreover, our results are confirmed when MIND is tested in Domain Incremental (DI) scenario (i.e. the same classes are presented for each new task but the context of the inputs changes). 

\end{itemize}

\section{Related Works}
Continual learning has gained significant attention in recent years and various approaches have been proposed to address the problem of catastrophic forgetting. This section provides a small overview of the most prominent works, however, more detailed descriptions of the field have been proposed in reviews such as \cite{survey_masana} and \cite{survey2}. These categorizations are not mutually exclusive, and many approaches may incorporate techniques from multiple categories. 

\paragraph{Architectural-Based}

Architectural-based approaches aim to modify the model architecture to alleviate catastrophic forgetting. The first works falling in this category are Progressive Neural Networks (PNN) \cite{pnn}, where the network is augmented with new connections spanning both height-wise and width-wise, and Dynamically Expandable Networks (DEN) \cite{DBLP:conf/iclr/YoonYLH18} where they cope with new tasks by splitting/duplicating units and timestamping them. Finally, PackNet, proposed by \cite{packnet} compresses several datasets into a single network and works as a multi-task architecture.

\paragraph{Regularization-Based}

Regularization-based approaches focus on modifying the learning objective or introducing regularization terms to preserve knowledge from previous tasks. Perhaps the most widely used method in this category is Learning without Forgetting (LwF) \cite{lwf}, which employs distillation to transfer knowledge from an old model to a new model that faces a new task. On the same line of work, where distillation is the major forgetting-preventing mechanism, Learning without Memorizing (LwM) \cite{lwm} proposes to distillate the attention heatmaps (obtained through Grad-CAM \cite{gradcam} ) to preserve previous spatial awareness of the model. Another common approach is to use a penalty term for each weight, as in Synaptic Intelligence (SI) \cite{si} and Memory Aware Synapses (MAS) \cite{mas}. Another seminal work is Elastic Weight Consolidation (EWC) \cite{ewc}, which penalizes weight changes that could disrupt previously learned knowledge through the computation of the fisher information at the end of each task. Finally, RWalk \cite{rwalk} builds upon EWC and introduces a KL-divergence-based Regularization on top of the standard methodology. Finally, PASS \cite{pass}, uses a prototype vector for each class to reduce catastrophic forgetting combined with a self-supervised learning technique to reduce task-level overfitting.

\paragraph{Rehearsal-Based}

Rehearsal-based approaches tackle catastrophic forgetting by explicitly storing and replaying past experiences during training. The first method proposed on this line is Experience Replay (ER) \cite{er}, where replay patterns of old tasks are randomly selected to be replayed in future tasks. Shortly after, a plethora of other methodologies have been proposed such as GDumb \cite{gdumb}, iCarl \cite{icarl} etc. 
Among the rehearsal-based approaches, we can devise a subsection called pseudo-rehearsal where replay data is generated through generative networks, towards this direction \cite{DBLP:conf/nips/ShinLKK17} constitutes the first work.

\begin{figure}[h!]
    \centering
    \includegraphics[width=0.95\columnwidth]{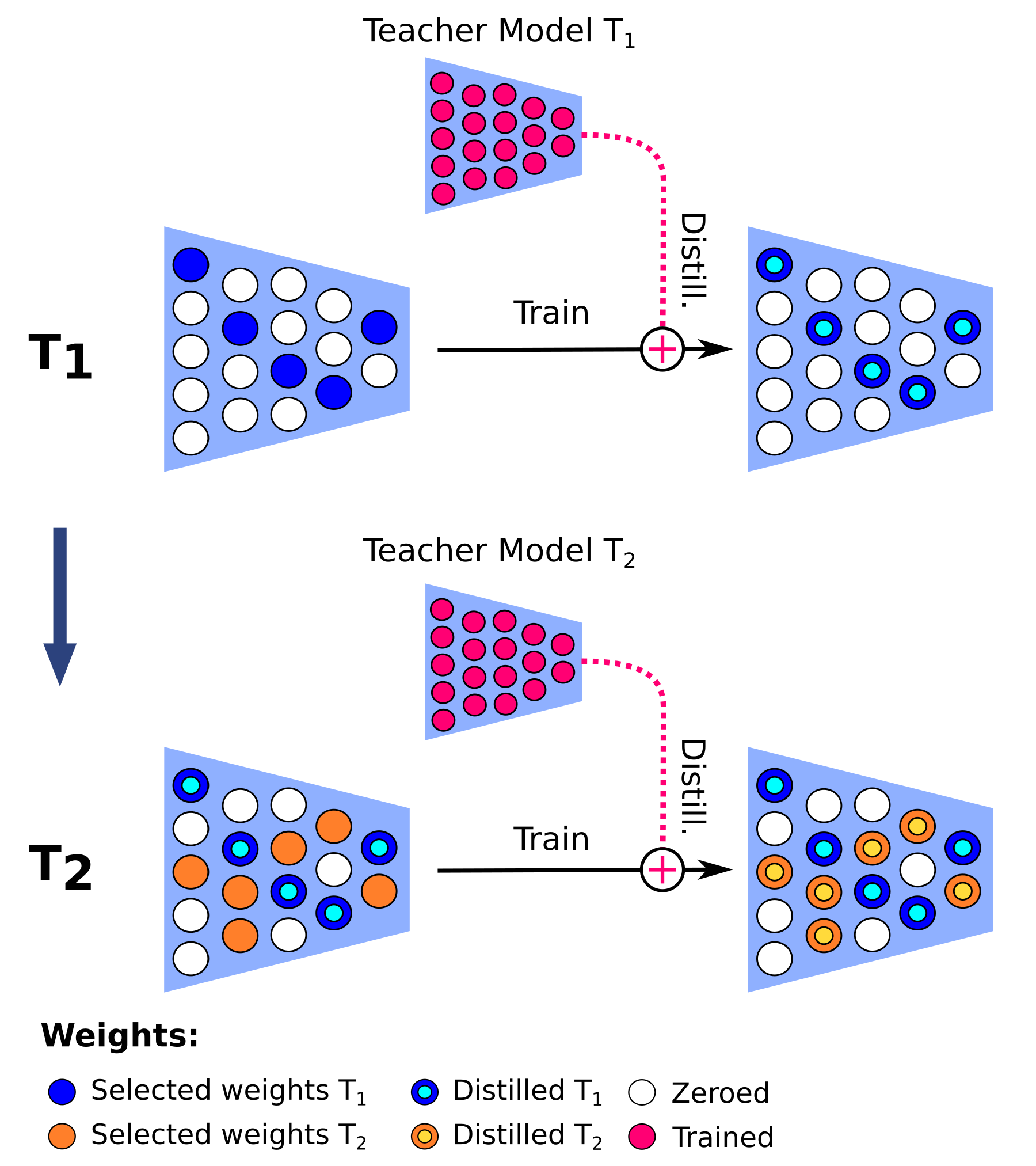}
    \caption{\emph{MIND training pipeline during Task 1 and Task 2.} After optimizing the new model (i.e. teacher model T$_1$), weights are selected randomly (i.e. blue circles) from MIND for distilling the knowledge from the teacher model (i.e. blue plus cyan circles).}
    \label{fig:1}
\end{figure}

\section{Method}

We consider a class incremental (CI) scenario, where data streams are split into separate N tasks $\mathcal{T}_i$ with $i = 0,...,N-1$. Each new task $\mathcal{T}_i$ is characterized by a set of data $\mathcal{X}_i$ and their respective labels $\mathcal{Y}_i$. The number of classes presented at each $\mathcal{T}_i$ is $\frac{M}{N}$ where $M$ is the total number of classes and these are not shared between the other tasks (i.e. $\mathcal{Y}_i \cap \mathcal{Y}_j = \varnothing$ if $i\neq j$). We focused on a replay-free CI scenario, where during the learning phase of $\mathcal{T}_i$ the access to samples of previous tasks is negated. During test time data stream $\mathcal{X}_{test}$ contains images from all the $M$ classes presented during $N$ tasks. 

MIND belongs to the category of parameter isolation approaches (\cite{packnet}, \cite{hat}),  which comprise sub-networks optimized for each specific task. To enhance the performance of those methods, we employ a distillation technique during the sub-network per-task finetuning process. This methodology eliminates the need for accessing past data, as all the knowledge about previously encountered classes is effectively retained within the MIND sub-networks.

In the subsequent sections, we will detail the various optimization procedures employed in MIND and demonstrate their application in a class incremental (CI) learning scenario.

\subsection{Sub-network Optimization}
\label{subs:sub_net_optim}

A commonly employed technique for optimizing sub-networks is illustrated in PackNet \cite{packnet}. In this method, the conventional network denoted as $f$ undergoes an iterative process of network pruning. This process selects specific learnable parameters that can effectively accommodate new tasks. This approach enables continual learning without requiring an increase in network capacity, while also minimizing the impact of performance degradation. The fundamental procedure involves three key steps: training the network $f$ on a specific task, pruning a certain fraction of its weights (i.e., setting them to zero), and subsequently retraining the network pruned ($\hat{f}$) to restore accuracy by accounting for the changes in network connectivity. The value of the free parameters related to the current task (i.e., those that are not set to zero) will be frozen in time once this finetuning step is over.

Once a new task is introduced to train $f$, all the weights are utilized during the forward pass for output computation. However, only the weights associated with the current task, excluding those about previous tasks, are optimized. This selective optimization process is followed by applying the aforementioned pruning and retraining strategies to finetune the network for the second task. This entire process is iteratively repeated until all the tasks have been completed. Importantly, all the knowledge acquired during previous tasks and stored in the frozen weights of each sub-network is always used as initialization knowledge for the new tasks (i.e. task 3 sub-network uses task 1 and 2 sub-networks weights when performing the forward pass).

In order to optimize $f$ and $\hat{f}$, Cross-Entropy loss $\mathcal{L}_{CE}$ eq. \ref{eq:CEloss} is applied in both training and re-training. 
\begin{equation}
    \mathcal{L}_{CE} = \sum_{i = 1}^{C} t_i log(p_i)
    \label{eq:CEloss}
\end{equation}
where $C$ is the number of classes in the current task, $t_i$ the true label and $p_i$ is the softmax probability of the $i^{th}$ class.

\subsection{MIND}
\label{MIND}
A crucial step for methods involving the sub-network optimization Sec. \ref{subs:sub_net_optim} implies re-training the pruned network $\hat{f}$ resulting in a reduced network capacity and consequent performance degradation.
%Hence, MIND incorporates a distillation mechanism into the optimization procedure to reduce the performance degradation caused by the network's reduced capacity.
Hence, MIND solves this issue by incorporating a distillation mechanism into the optimization procedure (Fig. \ref{fig:1}).
During each new task $\mathcal{T}_i$, a new network $g$ is initialized and trained from scratch on the new incoming task data ($\mathcal{X}_i$; $\mathcal{Y}_i$). Once $g$ is trained, it is used as the teacher model during the $\mathcal{T}_i$ distillation phase whereas we consider the network $f$ utilized in MIND as the student model. $f$ is iteratively pruned using a \emph{random policy} (RP) for weights selection. The RP involves randomly selecting a fraction of the available weights from the network $f$, where available weights refer to those weights that have not been chosen and optimized during the training of previous tasks. The knowledge from the freshly trained network $g$ is then distilled into the sub-network of $f$ corresponding to task $\mathcal{T}_i$.

To optimize the pruned network $\hat{f}$ during task $\mathcal{T}_i$ through distillation, we employ the Jensen-Shannon loss, denoted as $\mathcal{L}_{SD}$ in Eq. \ref{eq:SDloss}. At task $\mathcal{T}_i$, given the new network $g$ parameterized by weights $\psi$ and the pruned network $\hat{f}$ parameterized by weights $\phi_i$, we can define the distillation loss as follows:
\begin{equation}
\begin{aligned}
    \mathcal{L}_{SD} = \sum_{x_i \in X } & \frac{1}{2} D_{KL}(p(z|x,\psi)\parallel p(z_i|x_i,\phi_i)) +\\
    & \frac{1}{2}D_{KL}(p(z|x,\phi_i)\parallel p(z_i|x_i,\psi))
    \end{aligned}
    \label{eq:SDloss}
\end{equation}
where $D_{KL}$ is the Kullback-Leibler divergence and $p$ represents the softmax output of the logits $z_i$ of $f_\phi$ or $g_\psi$ given a batch input $X$. Importantly, during the back-propagation only the subset of the weights $\phi_i$ selected for the task $\mathcal{T}_i$ are updated while the rest is frozen to retain past acquired knowledge.

This distillation loss is combined with cross-entropy loss Eq. \ref{eq:CEloss}  using a hyperparameter $\beta$ as in Eq. \ref{eq:CE_SD_loss}.

\begin{equation}
\mathcal{L} = \mathcal{L}_{CE} + \beta \mathcal{L}_{SD}   
\label{eq:CE_SD_loss}
\end{equation}

\begin{figure*}[!b]
    \centering
    \includegraphics[width=.95\textwidth]{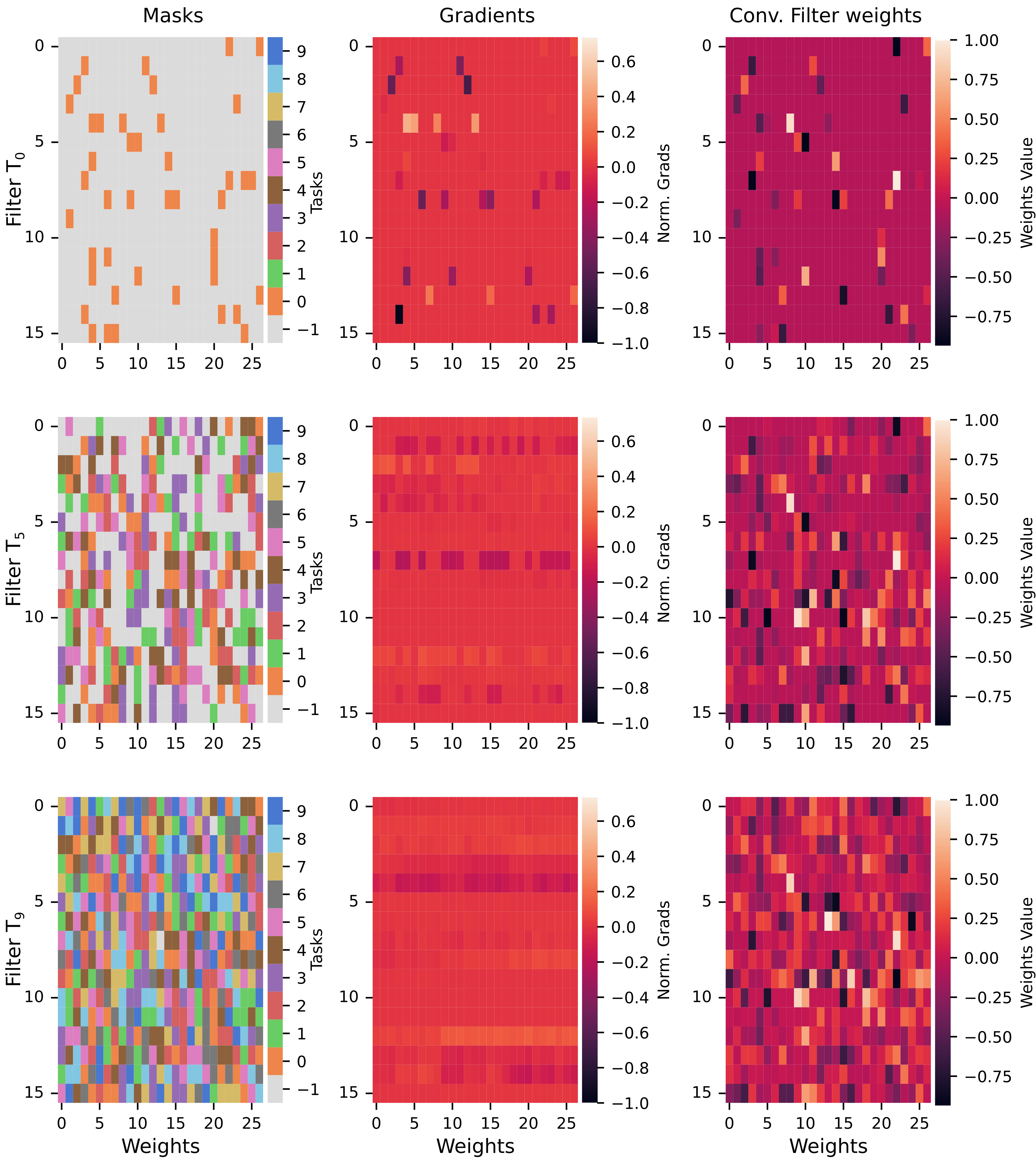}
    \caption{\emph{Visualization of masks, gradients, and weights on the first convolutional layer \texttt{conv\_1\_3x3} of the backbone}. Each row of the y-axis represents a single filter, unrolled on the x-axis. On the left side of the plot, the mask of the layer and their task assignment are reported; at the center, the gradient flow at the last batch; on the right, the values of the weights. On the top, we report the state after the first task, in the middle after the fifth, and at the bottom after the last task for CIFAR100/10.}
    \label{fig_sup:gradients}
\end{figure*}

\paragraph{Gating Mechanism}
We introduced a binary gating mask acting as a learning routing mechanism to guide the backpropagation procedure. This contribution redirects the flow of the gradient towards the active units of MIND (see Supp. Information Fig. \ref{fig_sup:gradients}).

Weights are defined as active (i.e. mask set to 1) if the parameters have been assigned to any sub-network, or as inactive if the parameters are not assigned yet (i.e. masks set to 0). Among active weights, old sub-network ones are frozen whereas current sub-network ones are updated during backpropagation.
By setting previous sub-networks weights to active, the current sub-network is learned by exploiting previously acquired knowledge (i.e. the forward computation takes into account also old sub-networks). This active/inactive masking procedure is also employed by each sub-network during the inference forward passes.

With this gating mechanism, the gradient computation is more precise and avoids discarding some of its magnitude that would be otherwise flowed towards inactive weights unlocking more fast and proficient learning. 

\paragraph{Batch Norm} To enhance the adaptability of MIND in CI and DI learning scenarios, we train the Batch-Norm layers~\cite{batchnorm} in each task and save the learned parameters corresponding to each sub-network. During the inference phase, we utilize the fitted Batch-Norm parameters that correspond to the selected sub-network. This contribution has been tested and proven to be highly effective in handling distributional shifts and achieving superior performance in our particular scenario, as demonstrated through ablation studies (Sec. \ref{sec:ablations}). This solution allows MIND to leverage task-specific Batch-Norm parameters, ensuring a better adaptation to each task and overcoming the limitations observed when Batch-Norm parameters are trained only during the initial task.
\begin{figure}[!hb]
    \centering
    \includegraphics[width=.95\columnwidth]{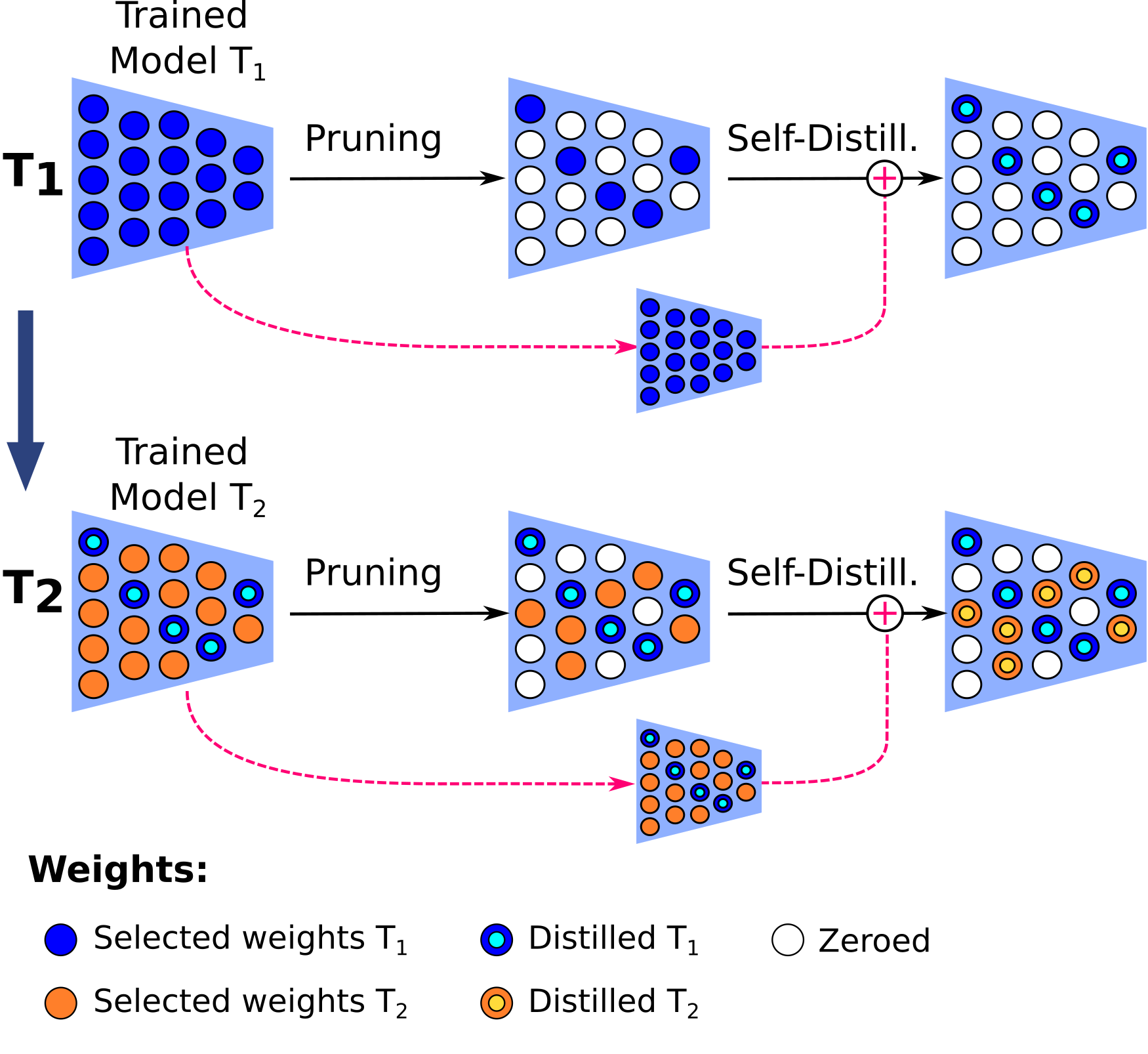}
    \caption{\emph{MIND training with self-distillation during task 1 and task 2}. After the optimization of all available weights of the model 
 during task $\mathcal{T}_i$, the most important weights policy is employed for selecting weights with the highest activation values (i.e. blue circles for $\mathcal{T}_1$) and pruning the remaining weights (white circles). These pruned weights serve as targets for distilling knowledge from the non-pruned network. Consequently, the distilled weights (i.e. blue and cyan circles for $\mathcal{T}_1$) are kept unchanged for the new incoming tasks.}  
    \label{fig:2}
\end{figure}

\subsection{MIND with Self-Distillation}
Our base distillation approach relies on initializing a new model for each new incoming task. However, in certain real-use case scenarios, hardware limitations such as in applications with low-power devices, impose constraints on available memory resources. For this reason, we explored a self-distillation procedure (Fig. \ref{fig:2}) which reduces the amount of memory used. 

Instead of using a new network reinitialized at each task, the free weights (zeroed in Fig. \ref{fig:2}) of MIND are directly trained on task $\mathcal{T}_i$. Then, we proceed with the pruning step and select the \emph{most important parameters} (MIP policy) trained on $\mathcal{T}_i$ which will be the target for our distillation (student sub-network). The distillation loss is the same as in Eq. \ref{eq:SDloss}, using MIND before pruning instead of the new model $g$. 

The MIP policy selects a fraction of weights with the highest absolute values for each layer. We assigned the same fraction of weights per task using all the available weights of the network (i.e., 10 $\%$ of weights in the scenario with 10 tasks). A depiction of the self-distillation procedure is presented in Fig.\ref{fig:2}

\subsection{Inference}
\label{subs:infer}
\begin{figure}[t!]
    \centering
    \includegraphics[width=.95\columnwidth]{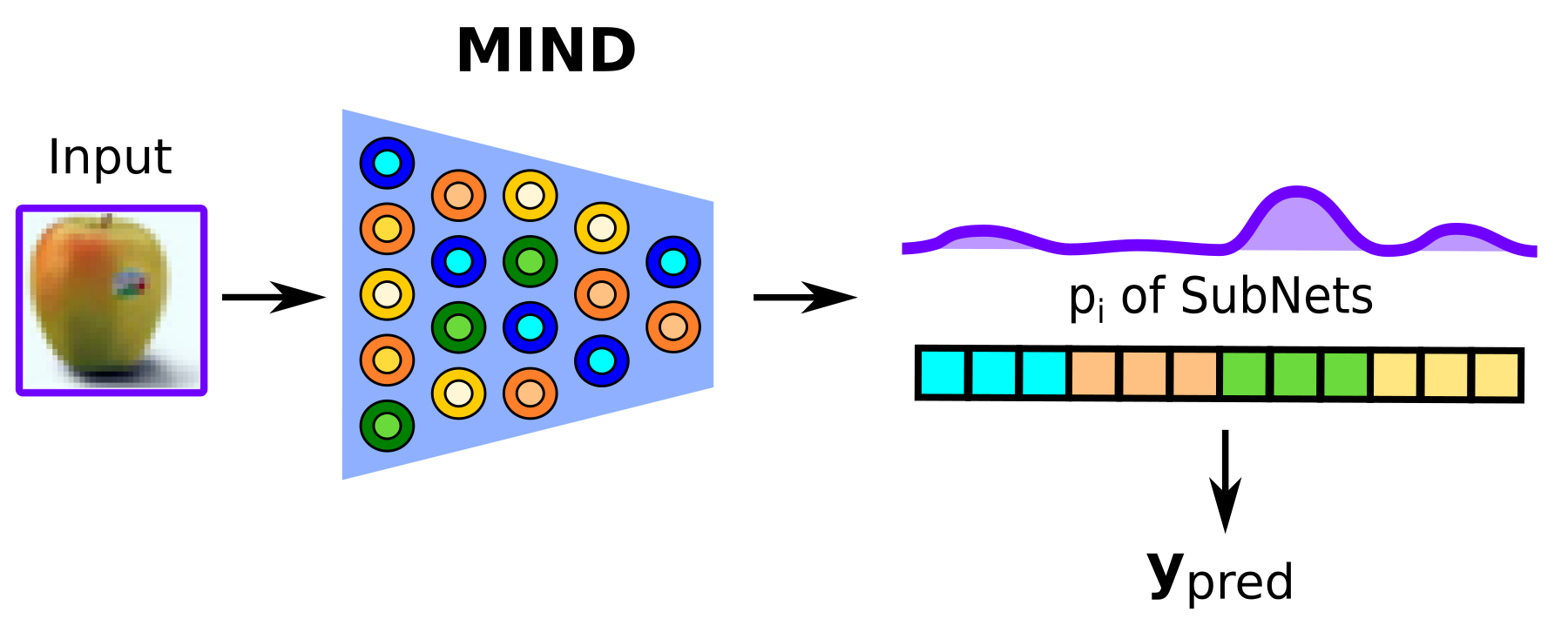}
    \caption{\emph{MIND inference overview}. For a given input image, MIND collects logit vectors from all sub-networks. During post hoc selection, the class with the highest probability computed from the logits vectors (Eq. \ref{eq:inference}), is selected.}    
    \label{fig:inference}
\end{figure}

During the inference phase (Fig. \ref{fig:inference}), each input image $x$ is fed through all the sub-networks of MIND, and the corresponding logits  $\textbf{z}_i$ vectors are collected (i.e. for each $\mathcal{T}_i$ there is a corresponding logits vector $\textbf{z}_i$). During inference, these sub-networks are retrieved through the binary active/inactive masking mechanism applied to the network weights described in Sec.~\ref{MIND}. After processing the input image through all the sub-networks, we compute the probability distributions $\textbf{p}_i$ from the softmax of the logits vectors $\textbf{z}_i$ scaled by a temperature $\tau$ (Eq. \ref{eq:inference}). 
\begin{equation}
\textbf{p}_i = \texttt{softmax}(\textbf{z}_i/\tau)
\label{eq:inference}
\end{equation}
From the distributions of probability $\textbf{p}_i$ with $i = 0,..., N-1$ where N is the number of tasks, we select as the predicted class the one with the highest likelihood. Through the \texttt{softmax} and temperature scaling~\cite{temp_scaling}, the logits vectors across sub-networks are respectively standardized and calibrated, obtaining comparable probability distributions of predictions.
\section{Experiments}
\begin{figure*}[h!]
    \centering
    \includegraphics[width=.95\textwidth]{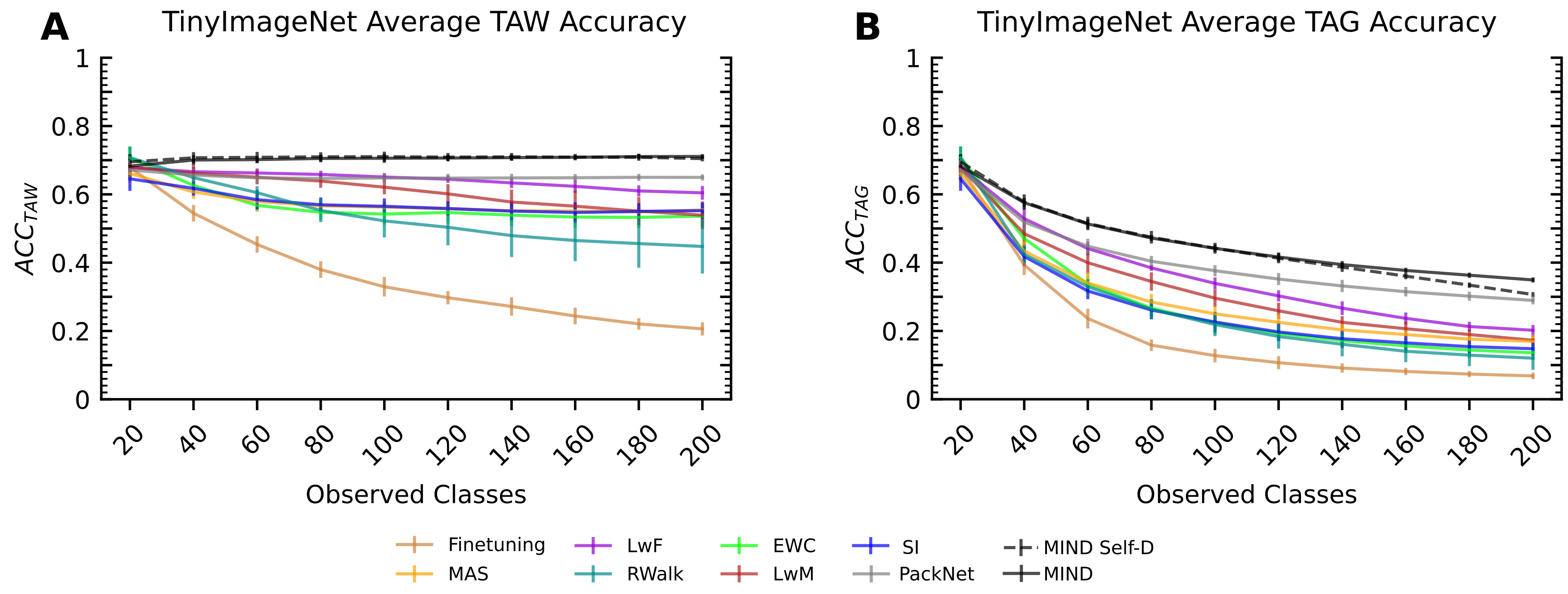}
    \caption{\emph{MIND outperforms all the state-of-the-art algorithms for CI learning}. \textbf{A-B}) Comparison between MIND and state-of-the-art CI learning algorithms ACC$_{TAG}$ on TinyImageNet/10 (A) and on Core50/10 (B). Results are reported as mean $\pm$ std across 10 runs obtained from 10 different seeds.}
    \label{fig:3}
\end{figure*}
For our experiments, we consider 4 datasets in the standard class-incremental (CI) learning scenario with all classes equally split among 10 tasks. More in detail we used: CIFAR100/10 \cite{cifar100} composed of $32\times32\times3$ images of 100 different classes split into 10 classes per task. TinyImageNet/10 \cite{tiny} composed of $64\times64\times3$ images with a total of 200 classes split into 20 classes per task. Core50/10 \cite{core50} composed of $64\times64\times3$ images of 50 domestic objects split into 5 classes per task. Synbols/10 \cite{synbols} composed of $64\times64\times3$ images of 200 ideograms of the Japanese alphabet split into 20 classes per task (see Supp. Information for details). We opted to use Synbols to facilitate future research to manipulate the latent space of the input distribution, unlocking a deeper understanding of the pros and cons of MIND.

We set the backbone of MIND to \texttt{gresnet32} a variation of a \texttt{resnet32} comprising the gating mechanism described in Sec.~\ref{MIND}. The dimension of the embeddings is set to $D = 64$ as all the competitors. The training hyper-parameters were optimized for each dataset: in brevity, we performed a grid search for each hyper-parameter on a subset of values empirically observed while training. Final numbers and more specs are reported in the Supp. Information. 

We report the task agnostic (no task-label) accuracy over all the classes of the dataset after training the last task:  $ACC_{TAG}=\frac{1}{C} \sum_{c=1}^{C} a_n$ where $C$ is the total number of classes in the dataset and $a_c$ the accuracy of the single class $c$. We also report the task aware setting, where at inference time we have access to the task label, unlocking the ability of MIND to query the correct sub-network: $ACC_{TAW}=\frac{1}{T} \sum_{t=1}^{T} a_t$ where $T$ is the total number of tasks and $a_t$ the accuracy on task $t$. 

We run the experiments on a machine equipped with: GPU NVIDIA GeForce RTX 3080, 11th Gen Intel(R) Core(TM) i9-11950H @ 2.60GHz processor, and 32 GB of RAM.

\begin{table*}[!htb]

    \begin{minipage}{.75\linewidth}
      \centering
      \small
      \begin{tabular}[t]{@{}clcllllllllll@{}}
%\toprule
%\multicolumn{5}{c}{\Large\textbf{Task-Agnostic Setting}} \\ % Super title row
\addlinespace
 \Large \textbf{A} &      &    \textbf{\# Params.}                 &\textbf{CIFAR100/10}   & \textbf{TinyImageNet}   & \textbf{Core50 (CI)}  &\textbf{Synbols}     \\\midrule \midrule

%\multicolumn{5}{c}{\Large\textbf{Task-Agnostic Setting}} \\ % Super title row
\addlinespace
\multirow{8}{*}{\rotatebox[origin=c]{90}{\textbf{Task-Aware}}}
& Finetuning        &   0.47M     & 38.3$^\ddagger$             & 20.6 $\pm$ 1.9           & 38.5 $\pm$ 7.0               & 42.8 $\pm$ 5.2    \\
&LwF                &   0.47M     & 76.6$^\ddagger$             & 60.4 $\pm$ 2.0           & 79.0 $\pm$ 5.1               & 93.8 $\pm$ 1.7    \\
&EWC                &   0.47M     & 56.7$^\ddagger$             & 53.6 $\pm$ 3.1           & 52.7 $\pm$ 6.7               & 83.8 $\pm$ 3.6    \\
&SI                 &   0.47M      & 53.1$^\ddagger$             & 55.2 $\pm$ 2.4           & 44.8 $\pm$ 8.2               & 81.5 $\pm$ 6.1    \\
&MAS                &   0.47M      & 58.6$^\ddagger$             & 55.4 $\pm$ 1.9           & 66.5 $\pm$ 3.5               & 84.7 $\pm$ 2.7    \\
&RWalk              &   0.47M      & 49.3$^\ddagger$             & 44.7 $\pm$ 7.9           & 40.4 $\pm$ 8.2               & 67.8 $\pm$ 8.3    \\
&LwM                &   0.47M      & 70.4$^\ddagger$             & 53.9 $\pm$ 4.0           & 67.2 $\pm$ 4.9               & 93.0 $\pm$ 1.8    \\
&PackNet            &   0.47M      & 72.4 $\pm$ 1.4       & 65.0 $\pm$ 0.9           & 95.7 $\pm$ 1.2               & 95.7 $\pm$ 1.1     \\
\rowcolor{black!5}
&\textbf{MIND (Self-D)} &  0.47M &\textbf{82.2 $\pm$ 0.5} &\textbf{70.7 $\pm$ 0.6}     &\textbf{99.8 $\pm$ 0.04}     &\textbf{98.7 $\pm$ 0.2} \\
\rowcolor{black!5}
&\textbf{MIND}    &  0.94M     &\textbf{82.3 $\pm$ 0.6} &\textbf{71.1 $\pm$ 0.7}     &\textbf{99.7 $\pm$ 0.08}     &\textbf{98.4 $\pm$ 0.2 } \\\bottomrule \bottomrule
\addlinespace
\multirow{10}{*}{\rotatebox[origin=c]{90}{\textbf{Task-Agnostic}}}
&Joint             &    0.47M      & 75.39$^\dagger$       & 59.38$^\dagger$          &94.8 $\pm$ 0.18         &99.4 $\pm$ 0.11      \\
&Finetuning        &    0.47M      & 10.1 $^\ddagger$             & 6.8 $\pm$ 0.9            &6.6 $\pm$ 2.8            &11.5 $\pm$ 4.0       \\
&LwF               &    0.47M      & 30.2 $^\ddagger$             & 20.2 $\pm$ 1.5           &15.0 $\pm$ 2.1         &47.3 $\pm$ 6.0       \\
&EWC               &    0.47M      & 13.1 $^\ddagger$             & 13.6 $\pm$ 2.2           &7.6 $\pm$ 2.3          &34.3 $\pm$ 4.7        \\
&SI                &    0.47M      & 13.6 $^\ddagger$             & 14.5 $\pm$ 2.0           &7.7$ \pm$ 1.2            &34.4 $\pm$ 6.9        \\
&MAS               &    0.47M      & 13.9 $^\ddagger$             & 16.9 $\pm$ 1.9            &11.0 $\pm$ 1.9           &29.7 $\pm$ 1.2      \\
&RWalk             &    0.47M      & 14.0 $^\ddagger$             & 12.0 $\pm$ 3.4            &7.1 $\pm$ 2.1            &23.2 $\pm$ 6.0       \\
&LwM               &    0.47M      & 21.9 $^\ddagger$             & 17.3 $\pm$ 1.5            &14.8 $\pm$ 1.5          &47.0 $\pm$ 4.6        \\
&PASS              &    11.2M      & 33.76 $^\dagger$      & 24.23 $^\dagger$         &-                      &-                   \\
&PackNet           &    0.47M      & 28.5 $\pm$ 2.2        & 29.0 $\pm$1.2            &40.0 $\pm$ 4.3          &60.4 $\pm$ 4.7        \\
\rowcolor{black!6}
&\textbf{MIND (Self-D)}& 0.47M &\textbf{35.7 $\pm$ 0.7} &\textbf{30.7 $\pm$ 0.7} &\textbf{55.9 $\pm$ 2.3} &\textbf{76.9 $\pm$ 1.0} \\
\rowcolor{black!6}
&\textbf{MIND}      &  0.94M    &\textbf{39.9 $\pm$ 0.9}  &\textbf{35.0 $\pm$ 0.8}     & \textbf{57.9 $\pm$ 2.1} &\textbf{76.5 $\pm$ 2.6} \\\bottomrule \bottomrule

%\bottomrule
\end{tabular}

% RESNET18 PARAMETRI 11 227 712
% RESNET32 PARAMETRI 469904 --> 470K
    \end{minipage}%
    \begin{minipage}{.25\linewidth}
        \centering
           \small
            
\begin{tabular}[t]{@{}ll@{}}
\Large \textbf{B}    & \textbf{Core50 (DI)} \\ \midrule 
LwF                     & 31.38 $\pm$ 0.02 \\
EWC                     & 27.91 $\pm$ 0.01 \\
SI                      & 25.5  $\pm$ 0.01 \\
\rowcolor{black!6}
\textbf{MIND}       & \textbf{79.28$\pm$2.63} \\ \bottomrule
\vspace{20mm}
\end{tabular}
            %\small
            %\input{tables/ablation_tab}
    \end{minipage} 
\caption{\textbf{A)} Comparison on CIFAR100/10, TinyImageNet/10, Core50/10, and Synbols/10 in Class-Incremental CI scenario with 10 tasks. All methods use \texttt{resnet32}. "Joint" here represents the case when the model is trained with all the classes available at once.  \textbf{B)} Comparison on Core50 Domain-Incremental (DI). ACC$_{TAW}$ and ACC$_{TAG}$ are reported as mean$\pm$ std across 10 runs obtained from 10 different seeds using Avalanche \cite{avalanche} framework and FACIL framework \cite{survey_masana} respectively. When std is not present the results are reported from literature. In particular, results marked by $^\ddagger$ are taken from the survey of \cite{survey_masana} while results marked by $^\dagger$  are taken from \cite{cotogni2022gated}.}
%\label{tab:results II}
\label{tab:resultsCI}
\end{table*}

\begin{figure}[t!]
    \centering
    \includegraphics[width=.95\columnwidth]{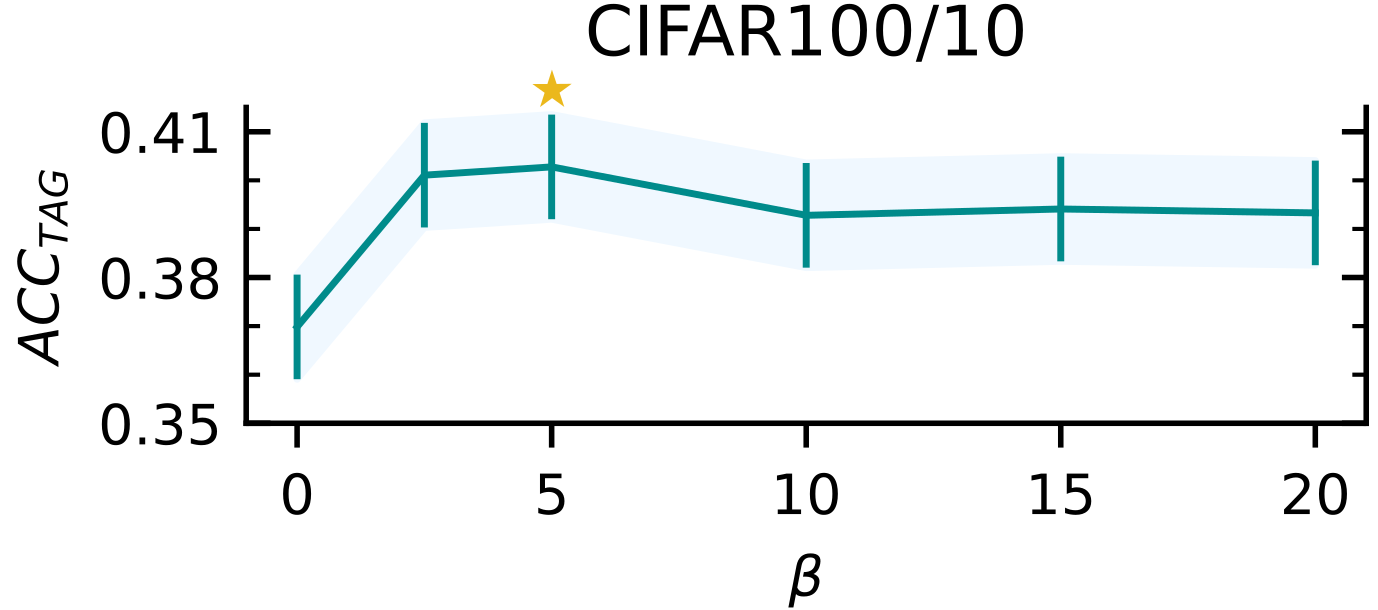}
    \caption{ACC$_{TAG}$ as a function of $\beta$ for the CIFAR100/10 dataset. Results are reported as mean $\pm$ std across 3 different runs from 3 different seeds. The star represents the final selected value.}
   
    \label{fig:4}
\end{figure}

\subsection{Class-Incremental and Domain-Incremental settings}
\label{sub_sec:evalCI}
The thorough assessment of MIND across diverse datasets encompassing CI and DI learning scenarios unveils consistent and reliable performances in both ACC$_{TAW}$ and ACC$_{TAG}$.

As it is evident from Tab. \ref{tab:resultsCI} A, our approach consistently outperforms all other methods in various benchmark settings. Notably, in the CIFAR100/10 dataset, MIND demonstrates a remarkable superiority, achieving approx. +6\% increase in $ACC_{TAG}$ and  +10\% increase in $ACC_{TAW}$ compared to the best existing memory-free technique documented in the literature. Particularly striking is our method's performance on the challenging TinyImageNet dataset, where it significantly outperforms all counterparts by a substantial margin in both $ACC_{TAG}$ and $ACC_{TAW}$ (approx. +6\% and +6\% respectively).

Analyzing Figure \ref{fig:3} B, it becomes evident that $ACC_{TAG}$ on the observed classes, while experiencing an initial decline, remains relatively steady thereafter. This observation underscores the effectiveness of accurate task identification after the initial tasks have been encountered (when the number of observed classes reaches or exceeds 80). This suggests a judicious balance between the plasticity and stability of MIND.

\begin{figure*}[!b]
    \centering
    \includegraphics[width=.95\textwidth]{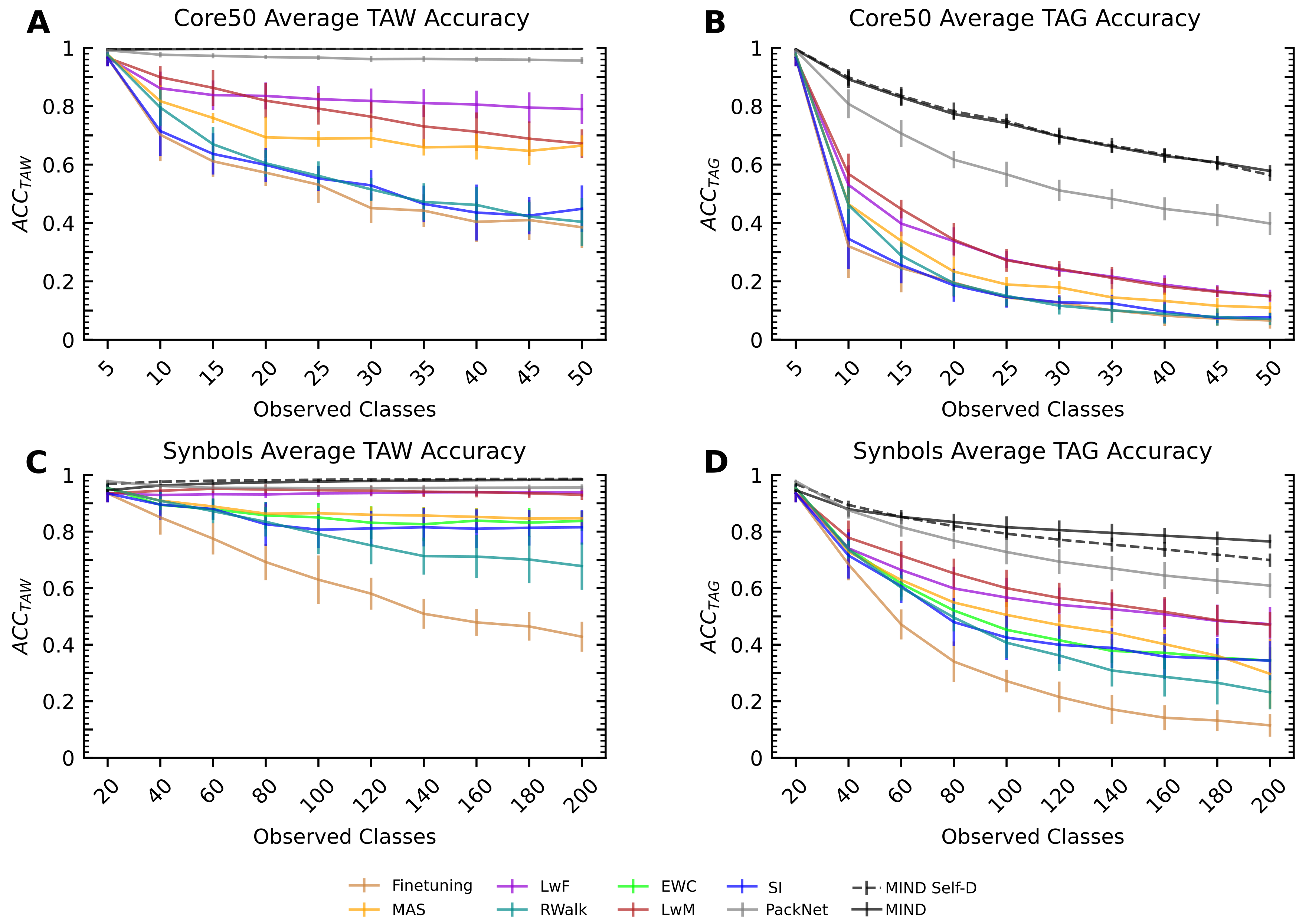}
    \caption{\emph{MIND outperform all the state-of-the-art algorithms for CI learning}. \textbf{A-B}) Comparison between MIND and state-of-the-art CI learning algorithms ACC$_{TAW}$ and ACC$_{TAG}$ on Core50/10, \textbf{C-D}) Comparison between MIND and state-of-the-art CI learning algorithms ACC$_{TAW}$ and ACC$_{TAG}$ on Synbols. Results are reported as mean $\pm$ std across 10 runs obtained from 10 different seeds.}
    \label{fig_sup:core50}
\end{figure*}

Importantly, these results are confirmed in Core50/10 and Synbols/10 datasets (Table \ref{tab:resultsCI} A and Supp. Information Fig. \ref{fig_sup:core50}). Collectively, these findings emphasize the significant progress made in the memory-free class-incremental scenario through the utilization of a multi-sub-networks paradigm with distillation, as exemplified by MIND.

To verify the efficacy of MIND compared to another parameter isolation method like PackNet, we conducted experiments on the 4 datasets described above for CI scenarios. The results obtained show that MIND consistently outperforms PackNet across different datasets (CIFAR100/10, TinyImageNet/10, Core50/10, Synbols/10) in ACC$_{TAW}$ and ACC$_{TAG}$. The improvements in ACC$_{TAW}$ highlight how our contributions increase the adaptability of each sub-network to novel tasks. Furthermore, the enhancements in ACC$_{TAG}$ underscore the superior task detection capabilities of MIND compared to PackNet across various sub-networks. This, in turn, leads to a more refined representation of the output probability distributions denoted as $\textbf{p}_i$ with $i = 0,...,N-1$, where $N$ represents the number of tasks.

For our experiments in Domain-Incremental (DI) scenarios, we consider the Core50 \cite{core50} dataset (for details see the Supp. Informations). We set the backbone of MIND to \texttt{gresnet18} with the dimension of the embeddings set to $D = 512$. We consider a DI learning scenario with 11 tasks where the same 50 classes are presented with a different background for each new task. From the results reported in Tab. \ref{tab:resultsCI} B, MIND more than doubles the DI learning results obtained by LwF, EWC, and SI. This result demonstrates how well our method copes with clear distribution shifts in input images thanks to which the proper sub-network configuration can be chosen easily during inference. 

\subsection{Self-Distillation}
Self-distillation in MIND represents a fundamental contribution that allows us to make use of the distillation mechanism and at the same time be compliant with common real use-case hardware limitations. Notably, our self-distillation approach yields task-aware results that are on par with those achieved using standard MIND. This alignment underscores the efficacy of our method in retaining task-specific information. Interestingly, the self-distillation technique maintains a remarkably low loss in ACC$_{TAG}$ if compared to standard MIND (Tab.~\ref{tab:resultsCI} A). This loss is significant only during the very last tasks (Fig.~\ref{fig:3} B) since the self-distillation becomes less efficient given the fact that the trainable parameters of the teacher model decrease as the number of tasks increases. Despite that, the self-distillation approach exhibits competitive performances over existing state-of-the-art methods in both ACC$_{TAW}$ and ACC$_{TAG}$ in all the datasets. 

Overall, our self-distillation technique stands as an optimal choice for hardware limitations contexts, offering a harmonious blend of efficient resource utilization and noteworthy performance outcomes. This makes it an attractive solution for systems with limited computational capabilities.

\begin{table}
\centering
    \begin{tabular}[t]{@{}ll@{}}
 \textbf{Ablation}    & \textbf{ACC$_{TAG}$} \\ \midrule 
\xmark ~Weight Sharing  & 37.2$\pm$0.8\\
\xmark ~Distillation    & 37.2$\pm$1.2\\
\xmark ~Batch-Norm    & 32.6$\pm$0.9\\ \bottomrule
\end{tabular}
    \caption{ Ablation studies ACC$_{TAG}$ results for CIFAR100/10. In each row of the table, we removed a component of MIND and reported the final ACC$_{TAG}$ result.}
    \label{tab:abl}
\end{table}

\section{Ablation Studies}
\label{sec:ablations}
Through the following ablation studies, we investigate the effects and contributions of the different components of MIND. All the results are reported for the CIFAR100/10 dataset, obtained through 5 different seeds for each experiment.

\paragraph{Weight Sharing}To evaluate the effectiveness of encapsulating a set of sub-networks into a single model, we conducted an ablation study by removing the weights-sharing between sub-networks. This experiment was crucial in identifying the advantages of using a cohesive set of sub-networks that incrementally share their weights, as opposed to an ensemble of independent sub-networks. The results of this experiment (Tab. \ref{tab:abl}) showed a decrease of $ACC_{TAG}$ of $-2.9\%$, demonstrating how the knowledge acquired during previous tasks is available for the new sub-networks and can also be used for increasing the current task knowledge.

\paragraph{Distillation}To assess the influence of the distillation loss, we conducted an ablation study by varying the parameter $\beta$ across the range [0-20]. The outcomes of this ablation investigation are graphically represented in Fig.~\ref{fig:4}. We observe a decrease of $-2.7\%$ when distillation is omitted ($\beta = 0$), as compared to the accuracy reported in Tab. \ref{tab:abl}. The distillation procedure plays a crucial role in effectively compressing the knowledge from new models and transferring it to sub-networks within MIND. Removing this component affects the overall performance, highlighting the importance of the distillation loss in achieving better accuracy and adaptation in the continual learning setup. To ensure coherence, we opted for a value of 5 for all other experiments, as it delivers the best performance when evaluated on CIFAR100/10.

\paragraph{Batch-normalization}A fundamental aspect to investigate in MIND is the effect of task-specific Batch-Norm parameters on the adaptability of the sub-networks on the CI learning scenario tasks. For this reason, we trained the Batch-Norm-layers only during the first task in both the new model and the sub-network $\mathcal{T}_1$ training and fixed them for the new incoming tasks. We observed a decrease in accuracy of $7.3\%$ (Tab. \ref{tab:abl}) which suggests how task-specific Batch-Norm parameters are highly effective in handling distributional shifts and achieving superior performances.

\section{Conclusion}
In this work, we introduced MIND, a rehearsal-free continual learning method. In particular, we proposed a new parameter isolation method that creates sub-networks tailored for each incremental task. MIND uses a distillation procedure to condense a new model (trained from scratch on each new task) in a sub-network of MIND, that will be exploited as compressed inter-knowledge thereafter. We also introduced a gating mechanism that optimizes the learning, guiding the gradient flow and selecting only the correct units that contributed during learning. This unlocks a more precise and reliable computation of the gradient providing more fast and proficient learning.  

Moreover, we proposed an alternative distillation procedure that can be used on systems with memory resource limitations. This alternative approach, called self-distillation, substitutes the role of the teacher (new model) during the distillation procedure with MIND itself.

Finally, we validated our results by running a wide batch of experiments encompassing 5 different benchmarks. Moreover, we ablated several architectural components of MIND and provided a sensitivity analysis of the distillation loss hyperparameter. Results show that MIND can be considered the new state-of-the-art method for class incremental rehearsal-free continual learning.
\section{Acknowledgments}
We would like to thank Marco Cotogni for his valuable suggestions on the manuscript and the reviewers for their detailed and valuable comments.

\bibliography{aaai24}
\section{Supplementary Information}
\subsection{Experimental Setup}
\subsubsection{MIND}
All our methods set the backbone as a \texttt{gresnet32} with masked outputs i.e. when a sub-network $i$ is queried, the output of the model is constrained to only the classes seen by that particular sub-network.
In Table \ref{tab:MIND_hyperp} we report the hyper-parameters used for MIND experiments divided into two main phases. First, the main phase where the fresh model is trained on a new task, and the distillation phase. All the hyperparameters have been kept the same for the self-distillation experiment too.

\subsubsection{Details on Class Incremental Benchmarks}
We compared MIND performances against state-of-the-art methods in the class incremental scenario for CIFAR100/10, TinyImageNet/10, Core50/10, and Synbols/10 datasets. We took advantage of the FACIL \textcolor{cyan}{https://github.com/mmasana/FACIL}\cite{survey_masana} library to perform experiments with the following models: Finetuning, LwF \cite{lwf}, EWC \cite{ewc}, SI \cite{si}, MAS \cite{mas}, RWalk \cite{rwalk} and LwM \cite{lwm}. For each dataset we set the model restnet32  as the network backbone and for a fair comparison with MIND we used the same augmentations (color jitter, random horizontal Flip, and random crop).
During the training with FACIL, we performed hyperparameter tuning using its grid search pipeline. The hyperparameter tuning is performed in the first 3 tasks and then the best hyperparameters found were fixed for the remaining tasks.

\subsection{Details on Domain Incremental}
To compare MIND against state-of-the-art methods in the domain incremental scenario, we took advantage of the library Avalanche \cite{avalanche}. All competitors' methods, namely: LwF \cite{lwf}, SI \cite{si} and EWC \cite{ewc} have been run with the aforementioned framework. 

To provide a fair comparison, we performed hyperparameter optimization for each method. In particular, we used the Optuna \cite{optuna} framework to maximize the final accuracy of the benchmark (\textcolor{cyan}{https://optuna.org}). For each method, the parameter space and the final parameters are reported in Table \ref{tab:hyperparam_di}. Each method's parameter space has been probed 10 times with the default optimization algorithm (i.e. Tree-structured Parzen Estimator). After choosing the appropriate hyperparameters for each method, we reported the final average accuracies across 10 different seeds.

\subsection{Details on Synbols}
\begin{figure*}[!h]
\begin{minipage}{0.30\textwidth}
    \centering
    \includegraphics[width=\textwidth]{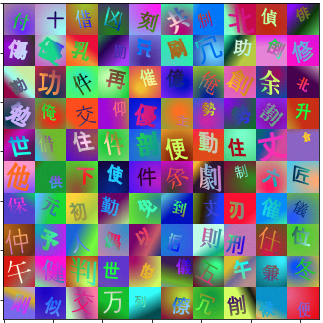}

\end{minipage}
\begin{minipage}{0.70\textwidth}
    
    \begin{verbatim}[
        
    # background creation
    bg = drawing.MultiGradient(alpha=0.9, 
                               n_gradients=1, 
                               types=('linear', 'radial'))
    # characters set
    alphabet = LANGUAGE_MAP["japanese"].get_alphabet(support_bold=True)
    
    # creation
    chars = alphabet.symbols[:200]
    fonts = alphabet.fonts
    basic_attribute_sampler(inverse_color=False, 
                            background = bg, 
                            char=lambda rng: rng.choice(chars), 
                            font=lambda rng: rng.choice(fonts))
    \end{verbatim}
    
\end{minipage}
\caption{Data depiction and code to generate the Synbols benchmark.}
\label{fig:synbols}
\end{figure*}

The Synbols dataset has been created using the code from the original repository (https://github.com/ServiceNow/synbols). The dataset consists of $32\times32\times3$ RGB images of chars. In Figure \ref{fig:synbols} we report the \texttt{python3} code to create the dataset along a plot of some data. The total number of characters (classes) is chosen to be 200 from the Japanese alphabet. We created 100k images for training and 10k for testing. While training we normalized the data by the statistics of the data (namely, mean and std for each channel) which we report to be: 
\begin{itemize}
    \item Train:\newline $\mu = [0.4792, 0.4790, 0.4782] \newline  \sigma = [0.2841, 0.2845, 0.2840]$
    \item Test:\newline $\mu = [0.4753, 0.4803, 0.4789]
    \newline\sigma = [0.2842, 0.2828, 0.2855]$
\end{itemize}

Along with the code to create a dataset with the same characteristic, a direct download to the created dataset is available at \textcolor{cyan}{https://github.com/Lsabetta/MIND}.

\begin{table*}[h!]
\centering
\begin{tabular}{@{}crlllll@{}}
 & \multicolumn{1}{l}{}                         & \textbf{CIFAR100/10} & \textbf{TinyImageNet} & \textbf{Core50 (CI)} & \textbf{Synbols} & \textbf{Core50 (DI)} \\ \midrule \midrule
 & \textit{bsize}                               & 256                  & *                     & *                    & *                & *                    \\
 & \textit{temp. $\tau$ $\ddagger$}                         & 6.5                  & 12                    & 3                    & 4                & 4                    \\\midrule

\addlinespace
\multirow{5}{*}{\rotatebox[origin=c]{90}{\textbf{Main Phase}}}
 
 & \textit{optim}                               & AdamW $^\dagger$     & *                     & *                    & *                & *                    \\
 & \textit{epochs}                              & 50                   & 100                   & 20                   & 25               & 15                   \\
 & \textit{lr}                                  & 0.005                & *                     & *                    & *                & 0.001                \\
 & \textit{milestones}                          & [35]                 & [70, 90]              & [15]                 & [10, 20]         & -                    \\
 & \textit{lr decay}                            & 0.5                  & *                     & *                    & *                & *                    \\ \midrule 
 
 \addlinespace
 \multirow{6}{*}{\rotatebox[origin=c]{90}{\textbf{Distill. Phase}}}
 & \textit{optim}                               & AdamW $^\dagger$     & *                     & *                    & *                & *                    \\
 & \textit{epochs}                              & 50                   & 120                   & 20                   & 25               & 20                   \\
 & \textit{lr}                                  & 0.035                & *                     & *                    & *                & 0.005                \\
 & \textit{milestones}                          & [40]                 & [80, 110]             & [15]                 & [15]             & -                    \\
 & \textit{lr decay}                            & 0.5                  & *                     & *                    & *                & *                    \\
 & distill. \textit{$\beta$}                    & 5                    & *                     & *                    & *                & *                    \\ \bottomrule \bottomrule
\end{tabular}
\caption{Hyperparameters governing the training of MIND across the array of distinct datasets. The symbol ``*'' signifies the adoption of the same values as those applied to the CIFAR100/10 dataset, while ``-'' means no usage. Note: $^\dagger$ the following params have been used $\beta s=(0.9, 0.999)$, $\epsilon = 10^{-08}$ and weight decay = 0. Note: $\ddagger$ the temperature has been tuned in a separate small validation set.}
\label{tab:MIND_hyperp}
\end{table*}

\begin{table*}[h!]
\centering
\begin{tabular}{@{}rllll@{}}
\multicolumn{2}{c}{\textbf{Shared Parameters for Grid Search}}              &  & \multicolumn{2}{c}{\textbf{Method Specific Params.}}              \\ \cmidrule(r){1-2} \cmidrule(l){4-5} 
\textit{lr first}     & (5e-1, 1e-1, 5e-2)             &  & \textit{LwF:}  & $\lambda_{lwf} = 10$, $T=2$       \\ \cmidrule(r){1-2} \cmidrule(l){4-5} 
\textit{lr}           & (1e-1, 5e-2, 1e-2, 5e-3, 1e-3) &  & \textit{LwM}   & $\beta = 2$, $\gamma_{lwm} = 1.0$ \\ \cmidrule(r){1-2} \cmidrule(l){4-5} 
\textit{lr searches}  & 3                              &  & \textit{EWC}   & $\lambda_{ewc} = 10000$           \\ \cmidrule(r){1-2} \cmidrule(l){4-5} 
\textit{lr min}       & 1e-4                           &  & \textit{MAS}   & $\lambda_{mas} = 400$             \\ \cmidrule(r){1-2} \cmidrule(l){4-5} 
\textit{lr factor}    & 3                              &  & \textit{SI}    & $\lambda_{si} = 10$              \\ \cmidrule(r){1-2} \cmidrule(l){4-5} 
\textit{lr patience:} & 10                             &  & \textit{RWalk} & $\lambda_{rwalk} = 20$              \\ \cmidrule(r){1-2} 
\textit{clipping}     & 10000                          &  &                 &                             \\ \cmidrule(r){1-2}
\textit{momentum}     & 0.9                            &  &                 &                             \\ \cmidrule(r){1-2}
\textit{wd}           & 0.0002                         &  &                 &                             \\
\end{tabular}

\caption{Hyperparameters used to run competitors' methodologies. The nomenclature follows the FACIL \cite{survey_masana} framework.}
\label{tab:MIND_hyperp}
\end{table*}

\begin{table*}[h!]
\centering
\begin{tabular}{@{}lll@{}}
    & \textbf{Search Space} & \textbf{Params. Chosen} \\ \midrule
\textbf{EWC} & $\eta \in [0.001, 0.007], \lambda_{EWC} \in [0.1, 0.5]$ &   $\lambda_{EWC}=0.375, \eta=0.0022$   \\ \midrule
\textbf{LwF} & $\eta \in [0.001, 0.007], \alpha_{LwF} \in [1, 5]$ &  $\alpha_{LwF}=4.073, \eta=0.0010$    \\ \midrule
\textbf{SI}  & $\eta \in [0.001, 0.007], \lambda_{si} \in [1, 5]$ &  $\lambda_{si}=2.019, \eta=0.0038$   \\ \bottomrule
\end{tabular}
\caption{Hyperparameters space probed 10 times with Tree-structured Parzen Estimator optimization algorithm and final hyperparams for each competitor. $\eta$ stands for learning rate. While $\lambda$ for SI has been kept fixed for all tasks.}
\label{tab:hyperparam_di}
\end{table*}

\begin{figure*}[!b]
    \centering
    \includegraphics[width=.95\textwidth]{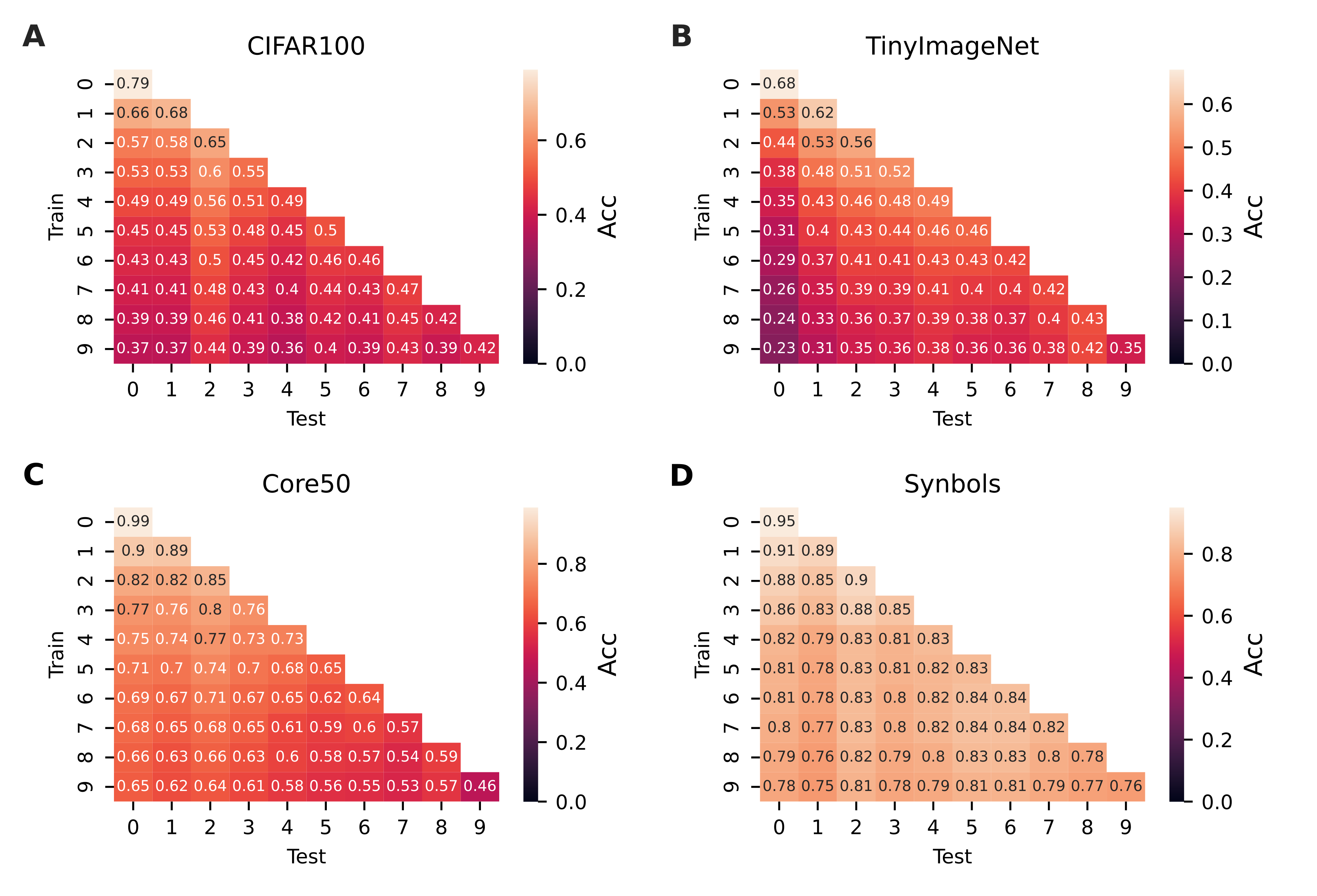}
    \caption{\emph{MIND $ACC_{TAG}$ performances during all tasks in diffrent dataset.} A-B-C-D) Matrices representing the task-agnostic accuracies for CIFAR100/10, TinyImageNet/10, Core50/10, and Synbols/10. The rows indices represent the task id. on which MIND has been trained whereas the columns indices identify the task id. of the test subset. Results are reported as mean $\pm$ std across 10 runs obtained from 10 different seeds.}
    \label{fig_sup:heatmaps}
\end{figure*}

\end{document}